\documentclass[10pt,twocolumn,letterpaper]{article}

\usepackage[pagenumbers]{cvpr} 
\usepackage{booktabs}

\usepackage{multirow}

\usepackage[dvipsnames]{xcolor}

\usepackage{soul}

\definecolor{cvprblue}{rgb}{0.21,0.49,0.74}
\usepackage[pagebackref,breaklinks,colorlinks,citecolor=cvprblue]{hyperref}

\def\paperID{5666} 
\def\confName{CVPR}
\def\confYear{2024}

\title{NeuSG: Neural Implicit Surface Reconstruction with 3D Gaussian Splatting Guidance}

\author{%
    Hanlin Chen
    \quad Chen Li
    \quad Yunsong Wnag
    \quad Gim Hee Lee
    \\
    Department of Computer Science, National University of Singapore \\ 
    \texttt{\{hanlin.chen, gimhee.lee\}@comp.nus.edu.sg}\\
}

\begin{document}
\maketitle
	\begin{abstract}
            
            Existing neural implicit surface reconstruction methods have achieved impressive performance in multi-view 3D reconstruction by leveraging explicit geometry priors such as depth maps or point clouds as regularization. However, the reconstruction results still lack fine details because of the over-smoothed depth map or sparse point cloud. In this work, we propose a neural implicit surface reconstruction pipeline with guidance from 3D Gaussian Splatting to recover highly detailed surfaces. The advantage of 3D Gaussian Splatting is that it can generate dense point clouds with detailed structure. Nonetheless, a naive adoption of 3D Gaussian Splatting can fail since the generated points are the centers of 3D Gaussians that do not necessarily lie on the surface. We thus introduce a scale regularizer to pull the centers close to the surface by enforcing the 3D Gaussians to be extremely thin. Moreover, we propose to refine the point cloud from 3D Gaussians Splatting with the normal priors from the surface predicted by neural implicit models instead of using a fixed set of points as guidance. Consequently, the quality of surface reconstruction improves from the guidance of the more accurate 3D Gaussian splatting. By jointly optimizing the 3D Gaussian Splatting and the neural implicit model, our approach benefits from both representations and generates complete surfaces with intricate details. Experiments on Tanks and Temples verify the effectiveness of our proposed method.
        
	\end{abstract}

         \begin{figure}[th]
          \centering
          \includegraphics[width=0.4\textwidth]{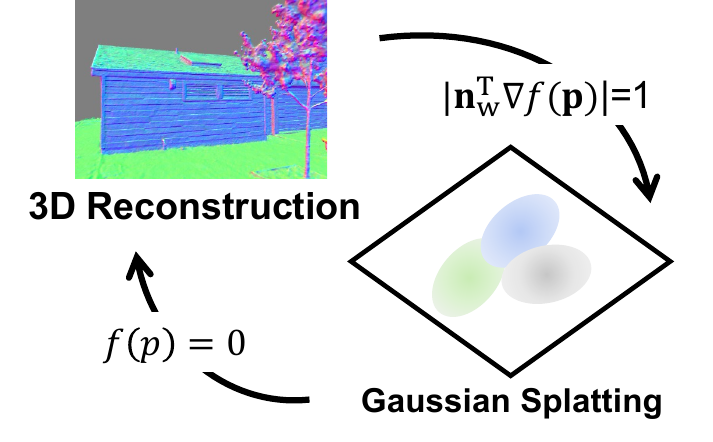} 
          \caption{An illustration of joint optimization of implicit surface reconstruction and 3D Gaussian Splatting.} \vspace{-2mm}
            \label{fig:core}
        \end{figure}
        
	\section{Introduction}
            Surface reconstruction from multiple calibrated views is a fundamental task in 3D computer vision. The conventional approach involves estimating a point cloud from images through multi-view stereo (MVS) techniques \cite{de1999poxels, broadhurst2001probabilistic, kutulakos2000theory, seitz1999photorealistic, seitz2006comparison}, followed by the extraction of a triangular mesh from the point cloud \cite{kazhdan2013screened, lorensen1998marching}. More recently, neural implicit surface reconstruction has emerged as a competitive alternative, particularly effective for surfaces with limited texture and non-Lambertian properties. These methods 
            use multi-layer perceptrons (MLP) or hash encoding \cite{muller2022instant} 
            to assign geometrical properties such as density \cite{nerf}, occupancy \cite{niemeyer2020differentiable}, or the signed distance to the nearest surface point \cite{wang2021neus, wang2023neus2, li2023neuralangelo} to spatial coordinates.
                
            The application of signed distance functions (SDF) in neural surface reconstruction is particularly noteworthy. It involves the use of an SDF-induced density function, allowing for volume rendering to learn an implicit SDF representation. While current neural implicit-based methods using rendering supervision alone yield impressive results for simple scenes, they face challenges with larger scenes, especially those containing extensive textureless areas \cite{yu2022monosdf, fu2022geo, zhang2022critical}. 
            Previous works have attempted to address this by integrating structural priors into the optimization process, such as depth prior \cite{niemeyer2022regnerf}, normal regularization \cite{yu2022monosdf}, point clouds regularization \cite{fu2022geo, zhang2022critical}, surface smoothness \cite{oechsle2021unisurf, zhang2021nerfactor}, or semantic regularization \cite{jain2021putting,guo2022neural}. While these methods succeed in generating complete surface reconstructions, they often result in over-smoothed surfaces lacking in fine details. Particularly, works such as those by \cite{fu2022geo, zhang2022critical} leverage sparse or dense point clouds from MVS to improve the fidelity of surface reconstruction. 
            Nevertheless, point clouds predicted from MVS are uniformly distributed and sometimes incomplete 
            and hence fails to provide geometry prior for some regions, especially regions with intricate details. Moreover, the generated points inevitably suffer from noisy geometries 
            that result in unreliable priors.

            The advantage of the 3D Gaussian Splatting \cite{kerbl3Dgaussians} is that it can generate dense point clouds with detailed geometry, 
            which provides sufficient geometric constraints especially for regions with intricate details. However, point cloud generated from the vanilla 3D Gaussian Splatting cannot be directly used as a prior since they are computed as the center of 3D Gaussians 
            that are generally located inside the surface. 
            To this end, we introduce a scale regularizer 
            that enforces the points to be close to the surface. Specifically, we enforce the smallest scaling factor of each 3D Gaussian ellipsoid to be close to zero such that the 3D Gaussian ellipsoid 
            are flattened to a plane. The extremely thin 3D Gaussian 
            are then inherently moved to the surface in order to render the correct color. Similar to existing point cloud-based approaches \cite{fu2022geo, zhang2022critical}, point clouds generated from 3D Gaussian Splatting are noisy 
            and therefore may provide incorrect prior. To 
            mitigate this problem, we further propose to refine the 3D Gaussians with the normal priors from the surface predicted by 
            a neural implicit model, NeuS \cite{wang2021neus}. Specifically, the direction of the smallest scaling factor can be regarded as the normal direction since the 3D Gaussian ellipsoid is close to a plane. We regularize the normal direction by aligning it with the normal predicted by NeuS. With this 
            joint optimization, 
            the point cloud generated from 3D Gaussians are
            refined to provide more reliable prior.
            As a consequence, our approach is able to benefit from both representations and generate a complete surface with intricate details. An illustration of our proposed mutual optimization is shown in Figure~\ref{fig:core}.

            In summary, our contributions include:
            
            \begin{itemize}
                \item We propose a novel framework that jointly optimizes NeuS and 3D Gaussian Splatting. This approach uses point clouds generated from 3D Gaussian Splatting to regulate NeuS, and 
                concurrently uses the predicted normals from NeuS to refine 3D Gaussian Splatting for higher quality point clouds.
                
                \item We introduce two regularizers to ensure that point clouds from extremely thin 3D Gaussians closely adhere to the surfaces. These include regularizing the smallest scaling of each 3D Gaussian close to zero and aligning the normals of these Gaussians perpendicular to the surfaces.
                
                \item We empirically demonstrate the effectiveness of our approach, dubbed NeuSG, highlighting its significant improvements over previous methods in surface reconstruction by experiments.
            \end{itemize}

	\section{Related Work}
            \subsection{Multi-view surface reconstruction}
                Surface reconstruction from multiple views is a fundamental aspect of 3D reconstruction within the field of computer vision. Traditional multi-view stereo (MVS) techniques \cite{bleyer2011patchmatch,de1999poxels,broadhurst2001probabilistic,kutulakos2000theory,schonberger2016pixelwise,seitz1999photorealistic,seitz2006comparison} utilize either feature matching for depth calculation \cite{bleyer2011patchmatch,schonberger2016pixelwise} or voxel-based shape representation \cite{de1999poxels,broadhurst2001probabilistic,kutulakos2000theory,seitz1999photorealistic,tulsiani2017multi}. The depth-based approaches focus on creating depth maps that are later combined into a comprehensive point cloud. In contrast, volumetric reconstruction methods estimate space occupancy and color within a voxel grid \cite{de1999poxels,broadhurst2001probabilistic,liu2020dist}, using color consistency checks to sidestep the need for explicit matching of correspondence. Nonetheless, the 
                finite resolution of the voxel grids limits the precision achievable by these reconstruction methods. Modern learning-based MVS approaches often modify traditional processes, such as feature matching \cite{luo2016efficient,ummenhofer2017demon,zagoruyko2015learning}, depth integration \cite{riegler2017octnetfusion}, or depth inference from multiple images \cite{huang2018deepmvs,yao2018mvsnet,zhang2020visibility,yao2019recurrent,yu2020fast}.
                
		\subsection{Neural Rendering and Radiance Fields}
                The pursuit of novel-view synthesis with volumetric representations began with the Soft3D \cite{penner2017soft}, which was later advanced by integrating deep-learning with volumetric ray-marching to form a continuous, differentiable density field for geometry representation \cite{henzler2019escaping,sitzmann2019deepvoxels}. Although volumetric ray-marching produces high-quality renders, its computational demand is substantial due to the extensive sampling required. The introduction of Neural Radiance Fields (NeRF) \cite{nerf} with importance sampling and positional encoding marked an improvement in render quality, albeit at the expense of speed due to the sizable Multi-Layer Perceptron (MLP).

                Subsequent methods have sought to optimize both quality and speed, often by implementing regularization strategies. To refine rendering quality, some works integrate position encoding or band-limited coordinate networks with neural radiance fields for a pre-filtered scene representation \cite{barron2021mip,barron2022mip,lindell2022bacon}. Different from these methods focusing on encodings and networks, S3IM \cite{xie2023s3im} introduces a multiplex training approach to further improve rendering fidelity. Recent innovations aim to expedite training and rendering by leveraging spatial data structures for feature storage, alternative encodings, and adjusting MLP size \cite{chen2022tensorf,fridovich2022plenoxels,garbin2021fastnerf,hedman2021baking,reiser2021kilonerf,takikawa2021neural}. Noteworthy techniques include the usage of hash tables for more efficient encoding and smaller MLPs or even bypassing MLPs entirely \cite{muller2022instant,fridovich2022plenoxels,sun2022direct}.

                Particularly, InstantNGP \cite{muller2022instant} employs a hash grid and occupancy grid for swift computation alongside a reduced MLP for depicting density and color, while Plenoxels \cite{fridovich2022plenoxels} utilizes a sparse voxel grid to interpolate a continuous density field, allowing it to eliminate neural networks entirely. Both methods incorporate Spherical Harmonics, with the former applying them to represent directional effects and the latter using them to encode color network inputs. Despite their impressive outcomes, these methods may face challenges in accurately representing empty space, influenced by the scene or capture conditions. Moreover, the structured grids chosen for computational acceleration can constrain image quality, and the intensive sampling required for ray-marching can slow down rendering. Addressing these limitations, the unstructured and explicit GPU-optimized 3D Gaussian Splatting \cite{kerbl3Dgaussians} presents a solution that achieves enhanced rendering speeds and improved quality without the reliance on neural components. In our paper, we utilize 3D Gaussian Splatting to generate dense point clouds closing the surface to obtain complete and detailed surfaces.

            \begin{figure*}[th]
              \centering
              \includegraphics[width=1\textwidth]{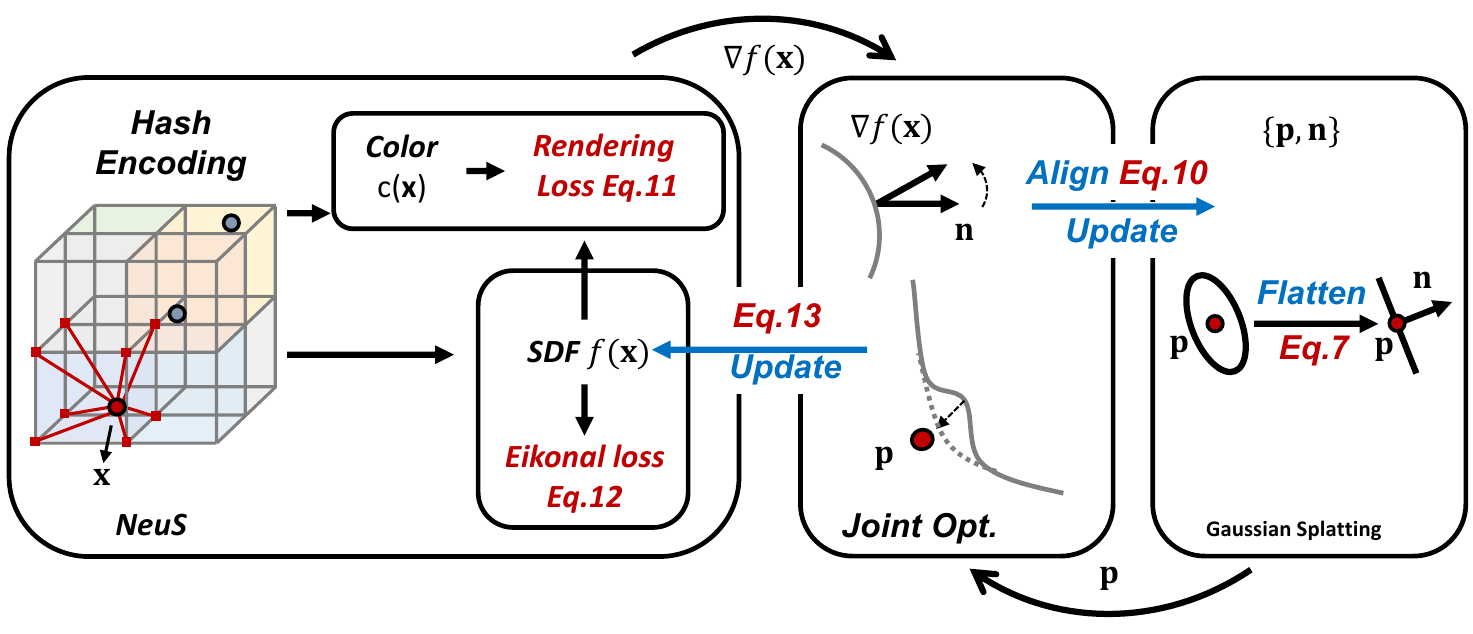} 
              \caption{The NeuSG framework includes three principal components: 1) Optimization of neural implicit surface reconstruction. 2) Geometric constraints from point clouds generated from Gaussian Splatting, as formalized in Eq.~\ref{eq:pt_loss}. 3) 
              Refinement of Gaussian Splatting through normal alignment, as detailed in Eq.~\ref{eq:align_normal}.}
                \label{fig:framework}
            \end{figure*}
		
		\subsection{Neural Surface Reconstruction}
                In 
                neural representation 
                of 3D surfaces, implicit functions such as occupancy grids \cite{niemeyer2020differentiable,oechsle2021unisurf} and Signed Distance Functions (SDFs) \cite{yariv2020multiview} have been preferred over basic volume density fields. To integrate these with neural volume rendering techniques \cite{nerf}, reparameterization methods 
                are developed to convert these representations back to volume density, as shown in previous research \cite{wang2021neus,yariv2021volume}. These neural implicit functions have proven more effective in surface prediction and maintaining the quality of view synthesis \cite{yariv2020multiview}.

                Further research has aimed to adapt these methods for real-time applications, 
                but compromising on 
                surface accuracy 
                as reported in recent studies \cite{li2022vox,wang2023neus2}. Simultaneously, additional research has considered extra data to enhance reconstruction quality \cite{darmon2022improving,fu2022geo,yu2022monosdf}. For example, NeuralWarp \cite{darmon2022improving} uses co-visibility cues from Structure-from-Motion (SfM) for patch warping, although this may not capture highly variable surfaces well. Other methods \cite{fu2022geo,zhang2022critical} apply sparse SfM point clouds to supervise the SDF 
                with their efficacy limited by the point cloud quality, as is the case with traditional techniques \cite{zhang2022critical}. The use of depth and segmentation data has been explored in studies using unrestricted image collections or hash-encoded scene representations \cite{sun2022neural,yu2022monosdf,zhao2022human}. While these can build complete geometry surfaces, they often lack sharpness and detail. Alternatively, some research \cite{wang2022hf} has introduced a multi-stage optimization to refine surface details using a displacement network to correct shapes from a primary network. At the same time, Neuralangelo \cite{li2023neuralangelo} employs hash encodings \cite{muller2022instant} and examines higher-order derivatives to reconstruct surfaces without auxiliary inputs. However, these approaches can result in imperfections 
                such as holes in less textured or observed areas.
                In comparison, our method utilizes point clouds from Gaussian Splatting to achieve completeness and detail in geometry.

	\section{Our Method}
            Our proposed NeuSG efficiently reconstructs complete and detailed surfaces of scenes from multi-view images. Section~\ref{sec:preliminary} provides an overview of NeuS \cite{wang2021neus} and 3D Gaussian Splatting \cite{kerbl3Dgaussians}. 
            Our proposed scale and normal regularization for 3D Gaussians is detailed in Section~\ref{sec:thin}. The joint optimization for both NeuS and 3D Gaussian splatting is 
            described in Section~\ref{sec:optim}. Figure~\ref{fig:framework} shows the illustration of our whole framework.

		\subsection{Preliminary}
                \label{sec:preliminary}
                \subsubsection{Neural Implicit Surfaces by Volume Rendering}
                    NeRF \cite{nerf} captures a 3D scene using density and color fields for synthesizing novel views through volume rendering. The method faces challenges in defining clear surfaces, often resulting in noisy and unrealistic surfaces when derived from density \cite{wang2021neus,yariv2021volume}. Signed Distance Functions (SDFs) offer a common method for representing surfaces implicitly as a zero-level set, $\{\mathbf{x} \in \mathbb{R}^3 \mid f(\mathbf{x}) = 0\}$, where $f(\mathbf{x})$ is the SDF value from an MLP $f(.)$. 
                    NeuS \cite{wang2021neus} replaces the volume density output of NeRF with SDF. A non-learnable but differentiable logistic function is then designed to convert SDF into opacity for volume rendering. Consequently, the intermediate SDF output allows for Eikonal regularization \cite{gropp2020implicit} to improve the quality of surface reconstruction.
                    Formally, the opacity $\alpha_i$ for a 3D point $\mathbf{x}_i$ with SDF value $f(\mathbf{x}_i)$ is given by:
                    \begin{equation}
                        \label{eq:opacity}
                        \alpha_i = \max \left( \frac{\Phi_s(f(\mathbf{x}_i)) - \Phi_s(f(\mathbf{x}_{i+1}))}{\Phi_s(f(\mathbf{x}_i))}, 0 \right),
                    \end{equation}
                    where $\Phi_s$ denotes the sigmoid function. 
                    Given a posed camera located at $\mathbf{o}$ and a ray direction $\mathbf{d}$, this opacity is then used to integrate the color radiance along a ray in the volume rendering process.
                    The color of each point $\mathbf{c}_i$ is predicted by an MLP, and the color of a pixel is the Riemann sum of these values:
                    \begin{equation}
                        \hat{\mathbf{C}}(\mathbf{o}, \mathbf{d}) = \sum_{i=1}^{N} w_i \mathbf{c}_i, \text{ where } w_i = T_i \alpha_i,
                    \end{equation}
                    where $\alpha_i$ is is the opacity defined in Eq.~\ref{eq:opacity}, and $T_i$ is the cumulative transmittance 
                    which indicates the fraction of light that reaches the camera. $N$ is the number of sample points along the ray. 

                \subsubsection{3D Gaussian Splatting}
                    As described in \cite{kerbl3Dgaussians}, 3D Gaussian Splatting is a method for representing 3D scenes with 3D Gaussians. Each Gaussian is defined by a covariance matrix $\boldsymbol{\Sigma}$, and a center point $\mathbf{p}$ 
                    which is the mean of the Gaussian. The 3D Gaussian distribution can be represented as:
                    \begin{equation}
                        \label{eq:gaussian_dist}
                        G(\mathbf{x}) = \exp{\{-\frac{1}{2}(\mathbf{x}-\mathbf{p})^\top \boldsymbol{\Sigma}^{-1}(\mathbf{x}-\mathbf{p})\}}.
                    \end{equation}
                    For optimization purposes, the covariance matrix $\boldsymbol{\Sigma}$ is expressed as the product of a scaling matrix $\mathbf{S}$ and a rotation matrix $\mathbf{R}$:
                    \begin{equation}
                        \boldsymbol{\Sigma} = \mathbf{R} \mathbf{S} \mathbf{S}^\top \mathbf{R}^\top.
                    \end{equation}
                    $\mathbf{S}$ is a diagonal matrix, stored by a scaling factor $\mathbf{s}$. The rotation matrix $\mathbf{R}$ is represented by a quaternion $\mathbf{r} \in \mathbb{R}^4$.

                    
                    For novel view rendering, the splatting technique \cite{yifan2019differentiable} is applied to the Gaussians on camera planes. Using the viewing transform $\mathbf{W}$ and the Jacobian of the affine approximation of the projective transformation $\mathbf{J}$ \cite{zwicker2001surface}, the transformed covariance matrix $\boldsymbol{\Sigma}'$ can be determined as:
                    \begin{equation}
                        \boldsymbol{\Sigma}' = \mathbf{J} \mathbf{W} \boldsymbol{\Sigma} \mathbf{W}^\top \mathbf{J}^\top.
                    \end{equation}
                    
                    A complete 3D Gaussian is 
                    given by its position $\mathbf{p} \in \mathbb{R}^3$, color represented with spherical harmonics coefficients $\mathbf{H} \in \mathbb{R}^k$, opacity $\alpha \in \mathbb{R}$, quaternion $\mathbf{r} \in \mathbb{R}^4$, and scaling factor $\mathbf{s} \in \mathbb{R}^3$. For a given pixel, the combined color and opacity from multiple Gaussians are weighted by Eq.~\ref{eq:gaussian_dist}. The color blending for overlapping points is:
                    \begin{equation}
                        \mathbf{C} = \sum_{i \in N} \mathbf{c}_i \alpha_i \prod_{j=1}^{i-1} (1 - \alpha_j),
                    \end{equation}
                    where $\mathbf{c}_i$, $\alpha_i$ denote the color and density of a point.

            \subsection{Regularization for 3D Gaussians} 
            \label{sec:thin}

                We propose 
                the utilization of 3D Gaussian Splatting to recover highly-detailed surfaces. This method is beneficial for creating dense point clouds that exhibit intricate details. However, the 3D Gaussian Splatting is not directly applicable in our context as the centers of the 3D Gaussians may not align with the actual surface. 
                To circumvent this problem, we introduce a scale regularization as described in Section~\ref{sec:min_scaling}. Furthermore, given that the points generated from 3D Gaussian Splatting are noisy, we propose to refine the 3D Gaussians using normal priors predicted by NeuS as discussed in Section~\ref{sec:normal}. The process is shown in Figure~\ref{fig:thin_gau}.

                \subsubsection{Scale Regularization} 
                    \label{sec:min_scaling}
                    As previously noted, conventional 3D Gaussian Splatting generates 
                    Gaussian centers $\{p_i\}$ that are typically positioned inside the surface 
                    and thus making 
                    them unsuitable for direct surface Regularization. To remedy this, we refine the 3D Gaussian ellipsoids into highly flat shapes. This process encourages the Gaussians to become significantly narrow, thereby pulling their centers closer to the surface. Within the 3D Gaussians, the scaling factor $\mathbf{s}$ defines the 
                    dimensions of the ellipsoid in each direction. By manipulating the scaling vector, we can alter the 
                    shape of the Gaussian. Specifically, we minimize the smallest component of the scaling factor $\mathbf{s} = (s_1, s_2, s_3)^\top \in \mathbb{R}^3$ for each Gaussian 
                    towards zero:
                \begin{figure}[ht]
                  \centering
                  \includegraphics[width=0.35\textwidth]{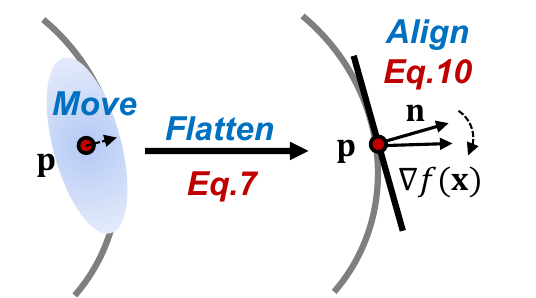} 
                  \caption{An 
                  illustration of} 
                  3D Gaussian flattening and normal alignment.
                    \label{fig:thin_gau}
                \end{figure}
                    \begin{equation}
                        \label{eq:min_scaling}
                        \mathcal{L}_{\text{s}} =  \|\min(s_1,s_2,s_3)\|_1.
                    \end{equation}
                    This process effectively flattens the 3D Gaussian 
                    and hence encourages the center points $\{p_i\}$ to align with the surface.

                \subsubsection{Normal Regularization} 
                    \label{sec:normal}
                    As the Gaussian flattens, the direction of the minimized scaling factor becomes the normal of the thin Gaussian. We define the normal in the camera coordinate system as:
                    \begin{equation}
                        \mathbf{n}_c = \operatorname{OneHot}\left(\arg\min(s_1,s_2,s_3)\right) \in \mathbb{R}^3,
                    \end{equation}
                    where $\operatorname{OneHot}(\cdot)$ converts an index to a vector. The normal $\mathbf{n}_c$ is then transformed to the world coordinate system using rotation $\mathbf{R}$:
                    \begin{equation}
                        \mathbf{n}_w = \mathbf{R} \times \mathbf{n}_c.
                    \end{equation}
                    We then align $\mathbf{n}_w$ with the surface normal predicted by NeuS to fine-tune the Gaussians:
                    \begin{equation}
                        \label{eq:align_normal}
                        \mathcal{L}_{\text{align}} =  \left|1 - | \mathbf{n}_w ^\top \cdot \nabla f(\mathbf{p}_i)| \right|_1.
                    \end{equation}
                    The absolute value ensures correct orientation 
                    regardless of the vector direction. 
                    %
                    The point cloud derived from 3D Gaussians is refined through the alignment of normals, thereby offering a more dependable prior for surface optimization.

		\subsection{Joint Optimization}
                \label{sec:optim}
                Initially, NeuS is optimized through a composite loss function 
                of rendering loss, Eikonal loss and the constraints imposed by point clouds derived from 3D Gaussians as described in 
                Section~\ref{sec:surf_optim}. 
                Subsequently, the normals as predicted by NeuS are utilized to further refine the 3D Gaussians, which is detailed in Section~\ref{sec:gau_optim}.
                
                \subsubsection{Implicit Surface Reconstruction}
                    \label{sec:surf_optim}
                    The 
                    training of the network is guided by a color loss that measures the discrepancy between the input images $\mathbf{C}$ and the rendered output $\hat{\mathbf{C}}$:
                    \begin{equation}
                        \mathcal{L}_{\text{RGB}} = \|\mathbf{C} - \hat{\mathbf{C}}\|_1.
                    \end{equation}
                    
                    To further refine the surface details, an Eikonal regularization term \cite{gropp2020implicit} enforces the correct gradient norm of the Signed Distance Function (SDF) in three-dimensional space:
                    \begin{equation}
                        \mathcal{L}_{\text{eik}} = \frac{1}{N} \sum_{i=1}^{N} \left( \|\nabla f(\mathbf{x}_i)\|_2 - 1 \right)^2,
                    \end{equation}
                    where $\nabla f(\mathbf{x}_i)$ is the gradient of the SDF $f(\mathbf{x}_i)$ evaluated at point $\mathbf{x}_i$. This is consistent with the NeuS framework for SDF-based volume rendering \cite{wang2021neus}.
                    
                    The point clouds $\mathbf{p}_i$ contribute to the surface optimization by enforcing the SDF values to be near zero at these points:
                    \begin{equation}
                        \mathcal{L}_{\text{pt}} = |f(\mathbf{p}_i)|_1,
                        \label{eq:pt_loss}
                    \end{equation}
                    where $f(\mathbf{p}_i)$ is the predicted SDF value at the point $\mathbf{p}_i$. The overall loss function combining these elements is:
                    
                    \begin{equation}
                        \mathcal{L}_{\text{total}} = \mathcal{L}_{\text{RGB}} + \lambda_1 \mathcal{L}_{\text{eik}} + \lambda_2 \mathcal{L}_{\text{pt}},
                    \end{equation}
                    with $\lambda_1$ and $\lambda_2$ balancing the individual components.
                    
                \subsubsection{Gaussian Splatting Refinement}
                    \label{sec:gau_optim}
                    In addition to the color matching loss $\mathcal{L}_{\text{RGB}}$, our approach involves minimizing the scaling factor through $\mathcal{L}_{\text{s}}$ as detailed in Eq.~\ref{eq:min_scaling}, and aligning the normals via $\mathcal{L}_{\text{align}}$ as per Eq.~\ref{eq:align_normal}. Hence, the combined loss function for Gaussian splatting is:
                    
                    \begin{equation}
                        \mathcal{L}_{\text{Gaussian}} = \mathcal{L}_{\text{RGB}} + \lambda_3 \mathcal{L}_{\text{s}} + \lambda_4 \mathcal{L}_{\text{align}},
                    \end{equation}
                    
                    where $\lambda_3$ and $\lambda_4$ weight the importance of the scaling and alignment terms, respectively. With this mutual optimization, the point cloud generated from 3D Gaussians will be refined, which in turn provides more reliable prior.

	\section{Experiment}
            We evaluate our method on the tasks of 3D surface reconstruction in Section~\ref{sec:comp}. Additionally, we validate the effectiveness of the proposed methods in Section~\ref{sec:ablation}.
            
                \begin{table*}[ht]
                \renewcommand{\arraystretch}{1.3} 
                \centering
                    \begin{tabular}{@{}c|ccccccc|c@{}}
                    \toprule
                            & Barn & Caterpillar & Courthouse & Ignatius & Meetingroom & Truck & Mean & GPU hours \\ \midrule
                    NeuralWarp \cite{darmon2022improving}  & 0.22 & 0.18        & 0.08       & 0.02     & 0.08        & 0.35  & 0.15 & -         \\
                    COLMAP \cite{schonberger2016structure} & 0.55 & 0.01        & 0.11       & 0.22     & 0.19        & 0.19  & 0.21 & -         \\
                    Vis-MVSNet \cite{zhang2020visibility}  & 0.49 & 0.21        & \textcolor{teal}{\textbf{0.36}}       & 0.25     & 0.43        & 0.28  & 0.34 & -         \\
                    NeuS \cite{wang2021neus}               & 0.29 & 0.29        & 0.17       & 0.83     & 0.24        & 0.45  & 0.38 & -         \\
                    NeuS-NGP                               & 0.46 & 0.32        & 0.08       & 0.81     & 0.08         & 0.44  & 0.37 & 16        \\
                    Geo-Neus \cite{fu2022geo}              & 0.33 & 0.26        & 0.12       & 0.72     & 0.20        & 0.45  & 0.35 & -         \\
                    MonoSDF \cite{yu2022monosdf}           & 0.49 & 0.31        & 0.12       & 0.78     & 0.23        & 0.42  & 0.39 & 18        \\
                    RegSDF \cite{zhang2022critical}        & -    & 0.22        & -          & -        & -           & 0.46  & -    & -         \\
                    NAngelo-19 \cite{li2023neuralangelo}   & 0.61 & 0.34        & 0.13       & 0.82     & 0.22        & 0.45  & 0.43 & 15        \\
                    NAngelo-22 \cite{li2023neuralangelo}   & \textcolor{cyan}{\textbf{0.70}} & \textcolor{cyan}{\textbf{0.36}}        & \textcolor{cyan}{\textbf{0.28}}       & \textcolor{teal}{\textbf{0.89}}     & \textcolor{cyan}{\textbf{0.32}}        & \textcolor{teal}{\textbf{0.48}}  & \textcolor{teal}{\textbf{0.50}} & 128       \\
                    NeuSG                                  & \textcolor{teal}{\textbf{0.73}} & \textcolor{teal}{\textbf{0.37}}        & 0.22       & \textcolor{cyan}{\textbf{0.83}}     & \textcolor{teal}{\textbf{0.35}}        & \textcolor{cyan}{\textbf{0.46}}  & \textcolor{cyan}{\textbf{0.49}} & 16        \\ \bottomrule
                    \end{tabular}
                    
                    \caption{
                    Quantitative assessment on the Tanks and Temples dataset \cite{knapitsch2017tanks} indicates that our} NeuSG achieves the highest quality in surface reconstruction. 
                    \textcolor{teal}{Teal} and \textcolor{cyan}{Cyan} indicate best and second best results, respectively.
                    Notably, NAngelo-19, representing NeuralAngelo configured with a maximum of $2^{19}$ hash entries per resolution (
                    same setting as our method), does not match NeuSG performance. 
                    NAngelo-22 which extends 
                    the capacity of NeuralAngelo to $2^{22}$ hash entries per resolution, necessitates tremendous increase in training duration but achieves results that are only comparable to 
                    our NeuSG. This highlights the efficiency of NeuSG, which attains similar or superior outcomes even with a less resource-intensive setup.
                    \label{tab:comp}
                \end{table*}

            \paragraph{Implementation Details.}    
            Our approach is implemented in PyTorch \cite{paszke2017automatic}, utilizing the Adam optimizer \cite{kingma2014adam} with a learning rate of \(1 \times 10^{-3}\) for the task of neural surface reconstruction. The weighting parameters \(\lambda_1\), \(\lambda_2\), \(\lambda_3\), and \(\lambda_4\) are set to 0.1, 1, 100, and 1, correspondingly. During training, we sample 1,024 rays at each iteration. Our method incorporates hash encoding, a multi-scale optimization strategy, and the numerical gradient techniques as proposed by NeuralAngelo \cite{li2023neuralangelo}. To enhance efficiency, we utilize \(2^{19}\) hash entries per resolution, which is less than the \(2^{22}\) used in NeuralAngelo, thus reducing both training time and memory usage while maintaining comparable results. The neural surface reconstruction model undergoes optimization over 500k iterations, interspersing the optimization of Gaussian Splatting every 100k iterations, totaling 30k iterations for the latter. In addition to point clouds derived from Gaussian Splatting, we integrate point clouds from Vis-MVSNet \cite{zhang2020visibility} for complement. Moreover, we adopt a dual-network architecture akin to NeRF++ \cite{zhang2020nerf++}, employing a NeRF model for external scene components and a NeuS model for internal geometry. The NeuSG is trained for about 16 hours on a single RTX4090 GPU with 24GB memory.

            \paragraph{Datasets.}
            Our experiments mainly focuses on large and complex scenes. We primarily evaluate our methodology on six scenes from the Tanks and Temples dataset \cite{knapitsch2017tanks}, encompassing a range of large-scale indoor and outdoor environments. These scenes comprise between 263 to 1,107 images 
            captured with a handheld monocular RGB camera. Ground truth data for evaluation is acquired via LiDAR sensors. Additional results are presented in the supplementary materials.

            \paragraph{Evaluation Criteria.}
            For the assessment of surface reconstruction, we report on the Chamfer distance and the F1 score \cite{knapitsch2017tanks}. Image synthesis quality is quantified using the Peak Signal-to-Noise Ratio (PSNR).

            \subsection{Comparisons to Baselines}
                \label{sec:comp}
\begin{figure*}[th]
                  \centering
                  \includegraphics[width=1\textwidth]{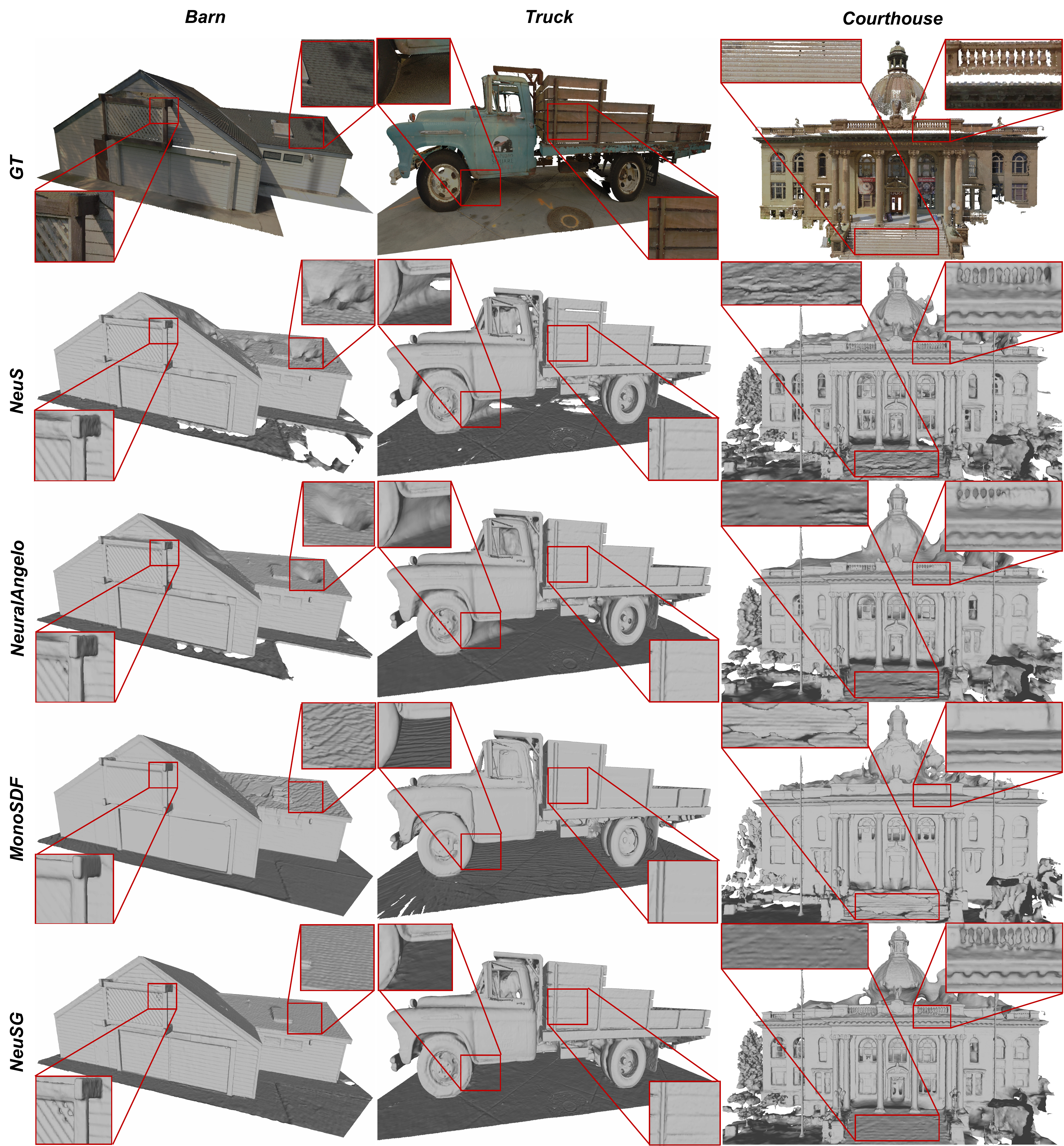} 
                  \caption{Qualitative comparison on Tanks and Temples dataset \cite{knapitsch2017tanks}. NeuSG excels in achieving both complete and intricately detailed surfaces, in contrast to baseline approaches which often result in surfaces that are either incomplete or marred by noise.}
                    \label{fig:comp}
                \end{figure*}

                We compare with state-of-the-art neural surface reconstruction methods, including NeuS \cite{wang2021neus}, NeuralWarp \cite{darmon2022improving}, Geo-Neus \cite{fu2022geo}, MonoSDF \cite{yu2022monosdf}, RegSDF \cite{zhang2022critical}, and NeuralAngelo \cite{li2023neuralangelo}. In addition, we have trained a variant of NeuS with hashing encoding, referred to as NeuS-NGP. For a fair comparison, all methods utilizing a hash table, such as NeuS-NGP, MonoSDF, NAngelo-19 and our NeuSG are constrained to a maximum of $2^{19}$ hash entries per resolution. NAngelo-19 denotes our own training of NeuralAngelo with the official code but limited to $2^{19}$ hash entries, whereas NAngelo-22 refers to the original implementation with $2^{22}$ hash entries as reported in their paper. The results for classical multi-view stereo including COLMAP \cite{schonberger2016structure} and Vis-MVSNet \cite{zhang2020visibility} are also presented for comprehensive evaluation. Qualitative comparisons are depicted in Figure~\ref{fig:comp} and quantitative results are summarized in Table~\ref{tab:comp}.

                We can see that NeuSG outperforms other approaches in terms of the F1 score. When compared to NeuS \cite{wang2021neus}, our method excels in reconstructing surfaces that are not only complete but also exhibit high-fidelity and intricate details (0.49 v.s. 0.38). 
                As depicted in Figure~\ref{fig:comp}, NeuSG shows its ability to recover full surfaces compared with NAngelo-19.
                Even when compared to NAngelo-22 \cite{li2023neuralangelo} 
                which utilizes a larger number of hash entries, our NeuSG still attains comparable performance (0.49 vs. 0.5) with significantly reduced computational time (16 hours vs. 128 hours). Furthermore, our method surpasses the geometric constraint-based approaches 
                \cite{yu2022monosdf,zhang2022critical,fu2022geo} 
                since they tend to produce over-smoothed results. This is particularly evident in Figure~\ref{fig:comp}, where MonoSDF produce overly smooth surfaces despite using geometric priors, whereas NeuSG preserves fine surface details. Compared with methods that also make use of point clouds for regularization 
                \cite{zhang2022critical,fu2022geo}, our method achieves better performance (0.49 v.s. 0.35) 
                and thus demonstrates the effectiveness of joint optimization of NeuS and Gaussian Splatting.

                As depicted in Figure~\ref{fig:comp}, our approach successfully achieves complete and detailed surface reconstruction. 
                For example, our method reconstructs the entire roof structure without any gaps in the `Barn' scene. This is a feat not matched by NeuS or NeuralAngelo, which tend to leave holes in the roof.
                Although MonoSDF is capable of reconstructing a complete roof, it falls short in preserving finer details 
                such as the 
                stripe patterns of the wall. Furthermore, our method avoids creating extraneous surface elements. A notable example is the `Truck' scene 
                where our method accurately reconstructs a standalone tire, in contrast to NeuS and NeuralAngelo which incorrectly merge the tire with the ground. 
                Additionally, our method distinctively captures the intricate details of the stairway and handrail in the `Courthouse' scene. This is a level of detail not replicated by the other methods.

            \subsection{Ablations}
                \label{sec:ablation}
                \subsubsection{Scale and Normal Regularization}
    

                    \begin{figure}[ht]
                        \centering
                        \begin{subfigure}{.25\textwidth}
                            \centering
                            \includegraphics[width=1\linewidth]{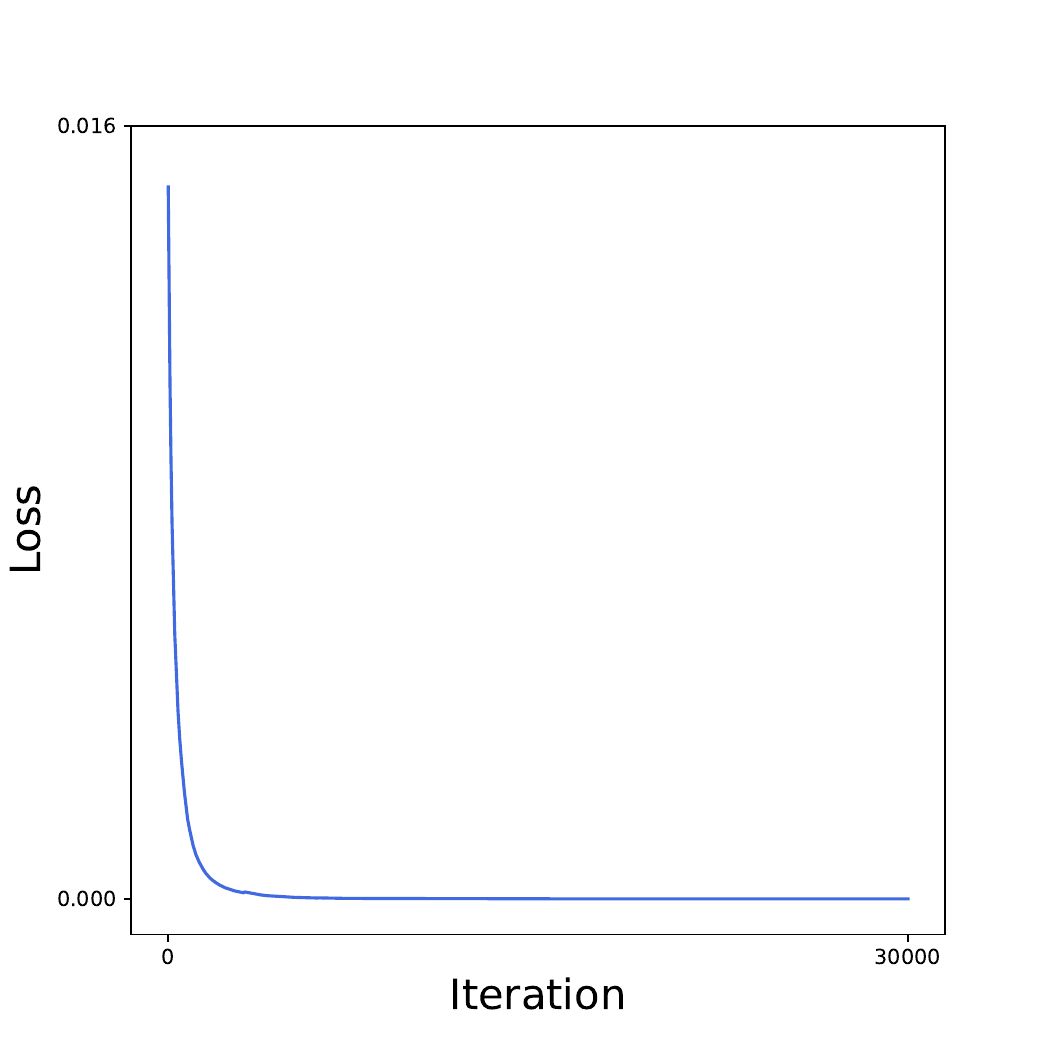}
                            \caption{Scale Regularization}
                            \label{fig:scale_reg}
                        \end{subfigure}%
                        \begin{subfigure}{.25\textwidth}
                            \centering
                            \includegraphics[width=1\linewidth]{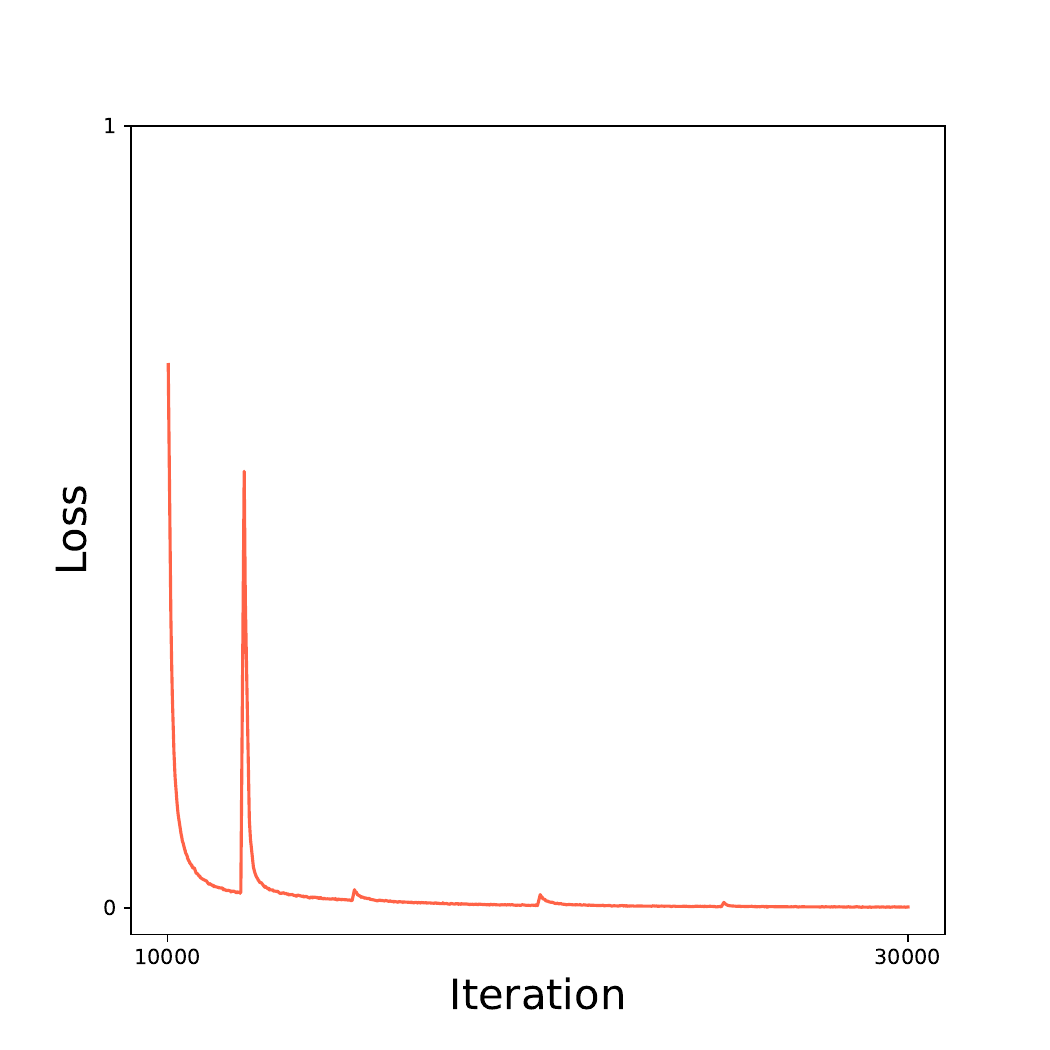}
                            \caption{Normal Regularization}
                            \label{fig:normal_reg}
                        \end{subfigure}
                        \caption{The scale regularizations for Gaussian Splatting. 
                        \textbf{Left:} Scale regularization. \textbf{Right:} Normal regularization.
                        }
                        \label{fig:reg}
                    \end{figure}

                    To validate the transformation of 3D Gaussians into extremely thin structures, we present the scale and normal regularization losses in Figure~\ref{fig:reg}. Figure~\ref{fig:scale_reg} demonstrates that the scale regularization loss approaches zero (approximately \(1 \times 10^{-8}\)), confirming the successful flattening of 3D Gaussians into extremely thin forms. Figure~\ref{fig:normal_reg} illustrates that the normal regularization loss also nears zero (about \(0.001\)), indicative of the accurate learning of normals.
                    
                    
                \subsubsection{Point Clouds Regularization}
                    \begin{table}[h]
                        \centering
                        \begin{tabular}{@{}cccc|c@{}}
                        \toprule
                        \multirow{2}{*}{MVS} & \multicolumn{3}{c|}{Gaussian} & \multirow{2}{*}{F1 Score} \\ \cmidrule(lr){2-4}
                                             & Original   & Scale  & Normal  &                           \\ \midrule
                        $\times$                    & $\times$          & $\times$      & $\times$       & 0.61 (+0)                 \\
                        \checkmark                    & $\times$          & $\times$      & $\times$       & 0.63 (+0.02)              \\
                        \checkmark                    & \checkmark          & $\times$      & $\times$       & 0.59 (-0.02)              \\
                        $\times$                    & $\times$          & \checkmark      & $\times$       & 0.69 (+0.08)              \\
                        $\times$                    & $\times$          & \checkmark      & \checkmark       & 0.73 (+0.12)              \\ \bottomrule
                        \end{tabular}
                        \caption{Ablations of different components.}
                        \label{tab:ablation}
                    \end{table}
                    
                    As depicted in Table~\ref{tab:ablation}, we conduct an ablation study to assess the impact of point cloud regularization in our method. The study commences with a baseline variant without point cloud regularization. Subsequently, we incorporate point clouds from Vis-MVSNet \cite{zhang2020visibility} to aid in the 
                    optimization process of the network. Following this, point clouds generated via Gaussian Splatting \cite{kerbl3Dgaussians} are employed to guide the learning of the Signed Distance Function (SDF). Our findings indicate that while the utilization of point clouds from Vis-MVSNet offers marginal improvements, the integration of point clouds from our proposed Gaussian Splatting method significantly enhances the 
                    performance of the model. 
                    
                    We also verify the effectiveness of scale and normal regularization in quality. We can see from the table 
                    that without any regularization, only point clouds from original Gaussian Splatting create a negative impact. By adding the scale and normal regularization, the performance is further improved 
                    to the best.

	\section{Limitation}
        Similar to numerous neural implicit reconstruction methods, our approach is constrained by the number of images available. Dense observations from multiple viewpoints are essential for accurate reconstruction of the object. This reliance on extensive multi-view data can be a significant limitation in scenarios where such comprehensive coverage is not feasible. Furthermore, our method, while effective in detailed surface reconstruction, may struggle in environments with sparse or uneven image distribution, potentially leading to less accurate reconstructions in these areas. 

        \section{Conclusion}
        In this work, we have introduced a neural implicit surface reconstruction pipeline 
        enhanced by the incorporation of 3D Gaussian Splatting 
        to facilitate the recovery of highly detailed and complete surfaces. The key advantage of 3D Gaussian Splatting lies in its capability to produce dense point clouds with detailed structures. 
        We 
        propose a scale regularization strategy that effectively transforms the 3D Gaussians into exceedingly thin shapes, thereby drawing the points closer to the surface. Furthermore, our method diverges from the conventional approach of using a static set of points as priors. Instead, we refine the 3D Gaussians using normal priors derived from surfaces predicted by neural implicit models. This approach endows 3D Gaussian Splatting with more precise directional guidance, leading to enhanced surface reconstruction. By jointly optimizing both the 3D Gaussian Splatting and the neural implicit model, our proposed method leverages the strengths of each technique, enabling the generation of complete surfaces with a high level of detail.
        
{
    \small

    \bibliographystyle{ieeenat_fullname}
    \bibliography{main}
}


\end{document}


\clearpage
\setcounter{page}{1}
\maketitlesupplementary

This supplementary material mainly includes: (1) More details about the experimental settings; (2) Additional Results.

\section{Details of experimental settings}
    Following prior work \cite{wang2021neus,yariv2021volume,li2023neuralangelo}, 
    our work operates under the assumption that the region of interest is confined within a unit sphere. The training process encompasses a total of $500,000$ iterations. We also incorporate the coarse-to-fine optimization strategy for hash encoding as introduced in \cite{li2023neuralangelo}. 
    The associated feature vectors are initialized to zero when a specific hash resolution is inactive. The learning rate is set at $1 \times 10^{-3}$, and it undergoes a linear warm-up phase over the first $5,000$ iterations. Subsequently, the learning rate is decreased by a factor of $10$ at two critical junctures: after $300,000$ and $400,000$ iterations. The optimization is carried out using the AdamW optimizer \cite{loshchilov2017decoupled} with a weight decay parameter set at $10^{-2}$. The parameter $w_{\text{eik}}$ is fixed at $0.1$. Furthermore, we use curvature regularization 
    which is a technique also utilized in \cite{li2023neuralangelo, zhang2022critical}. The weight of curvature regularization denoted as $w_{\text{curv}}$ is initially set at $5 \times 10^{-4}$, and is adjusted in accordance with the learning rate schedule. Regarding the neural network architecture, the Signed Distance Function (SDF) Multilayer Perceptron (MLP) comprises a single layer while the color MLP is constructed with four layers. 
    A batch size of $4$ is employed for experiments conducted on the Tanks and Temples dataset with the training being executed on a single GPU.

\section{Novel View Synthesis Results}
    
    In this work, approximately $7\%$ of images from the Tanks and Temples dataset are allocated for testing purposes while the remaining images are utilized for training. The results of this experimental setup are presented in Table~\ref{tab:psnr}--\ref{tab:lpips}. 
    We compare our methods with other neural reconstruction methods: NeuS \cite{wang2021neus}, MonoSDF \cite{yu2022monosdf}, and NeuralAngelo \cite{li2023neuralangelo}. As illustrated in the table, our methodology obtains the best performance on different evaluation metrics of PSNR, SSIM, and LPIPS compared with existing neural reconstruction methods.

    \begin{table}[h]
        \centering
        \begin{tabular}{@{}c|cccc@{}}
        \toprule
                    & NeuS & MonoSDF & NAngelo & NeuSG \\ \midrule
        Barn        & 23.11    & 23.40   & 22.90      & \textcolor{cyan}{\textbf{23.46}} \\
        Caterpillar & \textcolor{cyan}{\textbf{22.87}}    & 22.59   & 22.87      & 22.66 \\
        Courthouse  & 21.29    & 22.11   & 21.98      & \textcolor{cyan}{\textbf{22.18}} \\
        Ignatius    & 20.57    & 20.34   & 20.44      & \textcolor{cyan}{\textbf{20.60}} \\
        Meetingroom & 22.30    & 22.37   & 22.26      & \textcolor{cyan}{\textbf{22.78}} \\
        Truck       & 21.80    & 21.83   & \textcolor{cyan}{\textbf{21.94}}      & 21.81 \\ \midrule
        Mean        & 21.99    & 22.11   & 22.07      & \textcolor{cyan}{\textbf{22.25}} \\ \bottomrule
        \end{tabular}
        \caption{PSNR scores on the Tanks and Temples dataset \cite{knapitsch2017tanks}. \textcolor{cyan}{Cyan} indicate the best results.}
        \label{tab:psnr}
    \end{table}

    \begin{table}[]
        \centering
        \begin{tabular}{@{}c|cccc@{}}
        \toprule
                    & NeuS & MonoSDF & NAngelo & NeuSG \\ \midrule
        Barn        & 0.77     & 0.77    & 0.77       & \textcolor{cyan}{\textbf{0.78}}  \\
        Caterpillar & 0.71     & 0.70    & 0.71       & \textcolor{cyan}{\textbf{0.71}}  \\
        Courthouse  & 0.79     & 0.80    & 0.80       & \textcolor{cyan}{\textbf{0.81}}  \\
        Ignatius    & 0.58     & 0.57    & 0.58       & \textcolor{cyan}{\textbf{0.58}}  \\
        Meetingroom & 0.83     & 0.83    & 0.83       & \textcolor{cyan}{\textbf{0.84}}  \\
        Truck       & 0.77     & 0.77    & 0.77       & \textcolor{cyan}{\textbf{0.77}}  \\ \midrule
        Mean        & 0.74     & 0.74    & 0.74       & \textcolor{cyan}{\textbf{0.75}}  \\ \bottomrule
        \end{tabular}
        \caption{SSIM scores on the Tanks and Temples dataset \cite{knapitsch2017tanks}. \textcolor{cyan}{Cyan} indicate the best results.}
        \label{tab:ssim}
    \end{table}

    \begin{table}[]
        \centering
        \begin{tabular}{@{}c|cccc@{}}
        \toprule
                    & NeuS & MonoSDF & NAngelo & NeuSG \\ \midrule
        Barn        & 0.25     & 0.25    & 0.25       & \textcolor{cyan}{\textbf{0.23}}  \\
        Caterpillar & 0.26     & 0.27    & 0.26       & \textcolor{cyan}{\textbf{0.26}}  \\
        Courthouse  & 0.24     & 0.23    & 0.23       & \textcolor{cyan}{\textbf{0.22}}  \\
        Ignatius    & \textcolor{cyan}{\textbf{0.42}}     & 0.43    & 0.42       & 0.43  \\
        Meetingroom & 0.15     & 0.16    & 0.15       & \textcolor{cyan}{\textbf{0.15}}  \\
        Truck       & 0.21     & 0.21    & 0.22       & \textcolor{cyan}{\textbf{0.20}}  \\ \midrule
        Mean        & 0.25     & 0.26    & 0.25       & \textcolor{cyan}{\textbf{0.25}}  \\ \bottomrule
        \end{tabular}
        \caption{LPIPS scores on the Tanks and Temples dataset \cite{knapitsch2017tanks}. \textcolor{cyan}{Cyan} indicate the best results.}
        \label{tab:lpips}
    \end{table}

    \begin{figure*}[ht]
      \centering
      \includegraphics[width=1\textwidth]{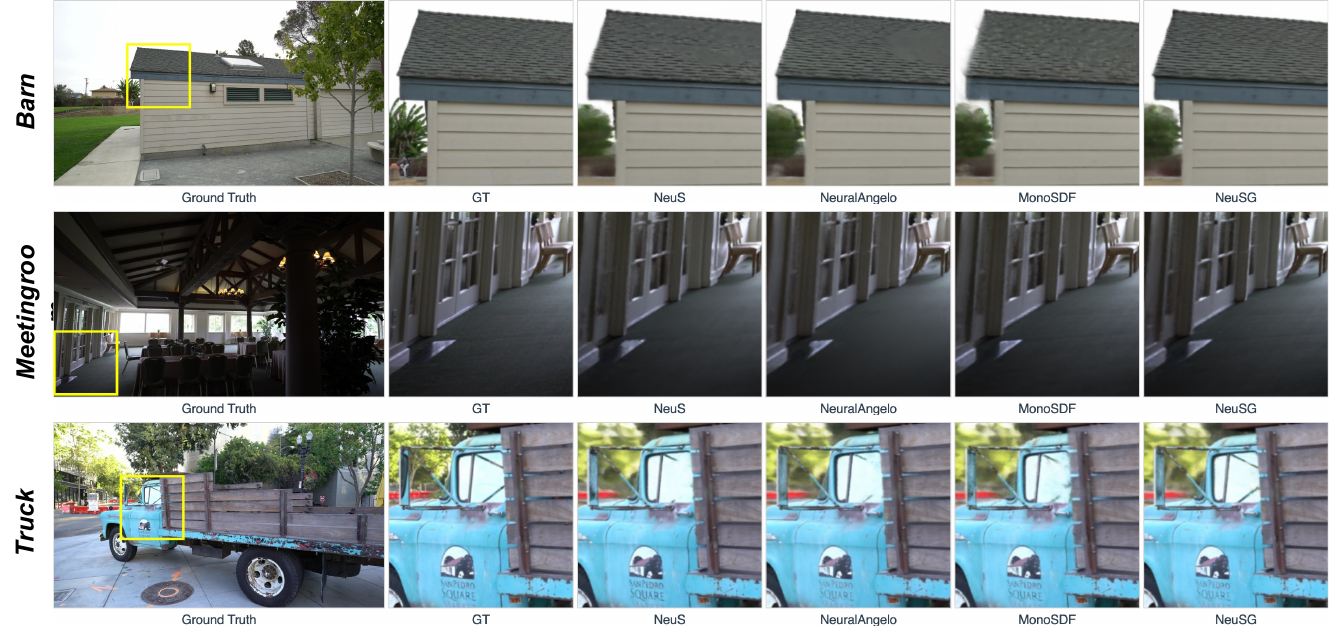} 
      \caption{Qualitative comparison on Tanks and Temples dataset \cite{knapitsch2017tanks}. NeuSG excels in achieving both complete and clear novel views synthesis, in contrast to baseline approaches which often result in synthesises that are either incomplete or blurry.}
        \label{fig:render}
    \end{figure*}
    
    Additionally, qualitative results are depicted in Figure~\ref{fig:render} 
    where our method achieves complete and clear novel views synthesis. For instance, it renders the roof in the `Barn' scene very well. This detail is not seen in the results from NeuS or NeuralAngelo, where the roof looks blurry. MonoSDF is able to recover the roof details, but it struggles with the roof edges. Our method also renders clearer images. A good example is in the `Meetingroom' scene 
    where our method shows a door more clearly than the others. Furthermore, it is good at rendering 
    reflections. 
    For example, our method shows the complete reflection on the mirror of the `Truck' scene that fails on other methods.

{
    \small
    \bibliographystyle{ieeenat_fullname}
    \bibliography{main}
}
